\documentclass[10pt,twocolumn,letterpaper]{article}

\usepackage{cvpr}
\usepackage{times}
\usepackage{epsfig}
\usepackage{graphicx}
\usepackage{amsmath}
\usepackage{amssymb}

\usepackage{caption}
\usepackage{multirow}
\usepackage{subcaption}

\usepackage[pagebackref=true,breaklinks=true,letterpaper=true,colorlinks,bookmarks=false]{hyperref}

\newcommand{\object}[1]{\texttt{#1}}
\newcommand{\predicate}[1]{\texttt{#1}}
\newcommand{\relationship}[3]{$\langle$\texttt{#1} - \texttt{#2} - \texttt{#3}$\rangle$}

\cvprfinalcopy 

\def\cvprPaperID{900} 

\begin{document}

\title{Action Genome: Actions as Composition of Spatio-temporal Scene Graphs}

\author{Jingwei Ji \quad Ranjay Krishna \quad Li Fei-Fei \quad Juan Carlos Niebles\\
Stanford University\\
{\tt\small \{jingweij, ranjaykrishna, feifeili, jniebles\}@cs.stanford.edu}
}

\maketitle

\begin{abstract}
Action recognition has typically treated actions and activities as monolithic events that occur in videos. However, there is evidence from Cognitive Science and Neuroscience that people actively encode activities into consistent hierarchical part structures. However in Computer Vision, few explorations on representations encoding event partonomies have been made. Inspired by evidence that the prototypical unit of an event is an action-object interaction, we introduce Action Genome, a representation that decomposes actions into spatio-temporal scene graphs. Action Genome captures changes between objects and their pairwise relationships while an action occurs.
It contains $10$K videos with $0.4$M objects and $1.7$M visual relationships annotated. With Action Genome, we extend an existing action recognition model by incorporating scene graphs as spatio-temporal feature banks to achieve better performance on the Charades dataset. Next, by decomposing and learning the temporal changes in visual relationships that result in an action, we demonstrate the utility of a hierarchical event decomposition by enabling few-shot action recognition, achieving $42.7\%$ mAP using as few as $10$ examples. Finally, we benchmark existing scene graph models on the new task of spatio-temporal scene graph prediction.
\end{abstract}


\section{Introduction}

Video understanding tasks, such as action recognition, have, for the most part, treated actions and activities as monolithic events~\cite{kinetics400,activitynet,hacs,charades}. Most models proposed have resorted to end-to-end predictions that produce a single label for a long sequence of a video~\cite{c3d,tsn,i3d,slowfast,timeception} and do not explicitly decompose events into a series of interactions between objects. On the other hand, image-based structured representations like scene graphs have cascaded improvements across multiple image tasks, including image captioning~\cite{anderson2016spice}, image retrieval~\cite{johnson2015image,schuster2015generating}, visual question answering~\cite{johnson2017inferring}, relationship modeling~\cite{krishna2018referring} and image generation~\cite{johnson2018image}. The scene graph representation, introduced in Visual Genome~\cite{krishna2017visual}, provides a scaffold that allows vision models to tackle complex inference tasks by breaking scenes into its corresponding objects and their visual relationships. However, decompositions for temporal events has not been explored much~\cite{lillo2014discriminative}, even though representing events with structured representations could lead to more accurate and grounded action understanding.

\begin{figure}[t]
    \centering
    \includegraphics[width=\linewidth]{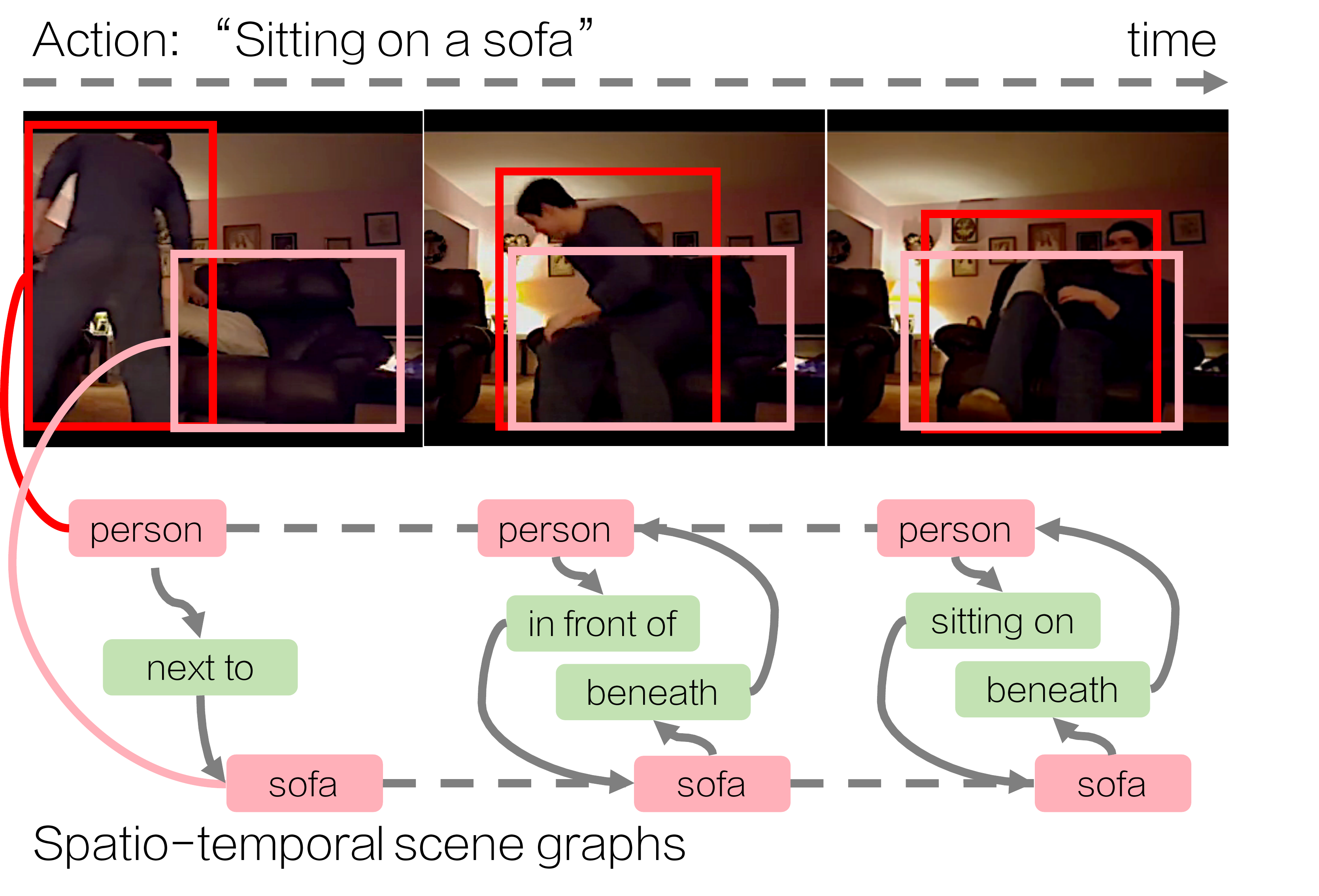}
    \caption{We present Action Genome: a representation that decomposes actions into spatio-temporal scene graphs. Inspired by hierarchical bias theory~\cite{zacks2001perceiving} and event segmentation theory~\cite{kurby2008segmentation}, Action Genome provides the scaffold to study the dynamics of actions as relationships between people and objects change. They also allow us to improve action recognition, enable few-shot action detection, and introduce spatio-temporal scene graph prediction.}
    \label{fig:pull}
\end{figure}

\begin{table*}[t]
\caption{A comparison of Action Genome with existing video datasets. Built upon Charades \cite{charades}, Action Genome is the first large-scale video database providing both action labels and spatio-temporal scene graph labels.}
\label{tab:dataset_compare}
\resizebox{\linewidth}{!}{%
\begin{tabular}{@{\extracolsep{4pt}}l rrr ccrr ccrr@{}}
 \multirow{2}{*}{Dataset} & Video & \# videos & \# action & \multicolumn{4}{c}{Objects} & \multicolumn{4}{c}{Relationships}  \\ 
 \cline{5-8}\cline{9-12}
 & hours & & categories & annotated & localized & \# categories & \# instances & annotated & localized & \# categories & \# instances \\ 
 \hline
ActivityNet~\cite{activitynet} & 648 & 28K & 200 &  &  & - & -&  &  & - & -\\
HACS Clips~\cite{hacs} & 833 & 0.4K & 200 &  &  & - & -&  &  & - & -\\
Kinetics-700~\cite{kinetics700} & 1794 & 650K & 700 &  &  & - & -&  &  & - & -\\ \hline
AVA~\cite{ava} & 108 & 504K & 80 &  &  & - & -& \checkmark &  & 49 & -\\
Charades~\cite{charades} & 82 & 10K & 157 & \checkmark &  & 37 & -&  &  & - & -\\
EPIC-Kitchen~\cite{epickitchen} & 55 & - & 125 & \checkmark &  & 331 & -&  &  & - & -\\
DALY~\cite{daly} & 31 & 8K & 10 & \checkmark & \checkmark & 41 & 3.6K &  &  & - & -\\
CAD120++~\cite{evar} & 0.57 & 0.5K & 10 & \checkmark & \checkmark & 13 & 64K & \checkmark & \checkmark & 6 & 32K\\
\hline
\textbf{Action Genome} & 82 & 10K & 157 & \checkmark & \checkmark & 35 & \textbf{0.4M} & \checkmark & \checkmark & 25 & \textbf{1.7M}
\end{tabular}}
\end{table*}

Meanwhile, in Cognitive Science and Neuroscience, it has been postulated that people segment events into consistent groups~\cite{michotte2017perception,barker1951one,barker1955midwest}. Furthermore, people actively encode those ongoing activities in  a hierarchical part structure --- a phenomenon referred to as hierarchical bias hypothesis~\cite{zacks2001perceiving} or event segmentation theory~\cite{kurby2008segmentation}. Let's consider the action of ``sitting on a sofa''. The person initially starts off next to the sofa, moves in front of it, and finally sits atop it. Such decompositions can enable machines to predict future and past scene graphs with objects and relationships as an action occurs: we can predict that the person is about to sit on the sofa when we see them move in front of it. Similarly, such decomposition can also enable machines to learn from few examples: we can recognize the same action when we see a different person move towards a different chair. While that was a relatively simple decomposition, other events like ``playing football'', with its multiple rules and actors, can involve multifaceted decompositions. So while such decompositions can provide the scaffolds to improve vision models, how is it possible to correctly create representative hierarchies for a wide variety of complex actions?

In this paper, we introduce Action Genome, a representation that decomposes actions into spatio-temporal scene graphs. Object detection faced a similar challenge of large variation within any object category. So, just as progress in 2D perception was catalyzed by taxonomies~\cite{miller1995wordnet}, partonomies~\cite{miller1976language}, and ontologies~\cite{yao2007introduction,krishna2017visual}, we aim to improve temporal understanding with Action Genome's partonomy. Going back to the example of ``person sitting on a sofa'', Action Genome breaks down such actions by annotating frames within that action with scene graphs. The graphs captures both the objects, \object{person} and \object{sofa}, and how their relationships evolve as the actions progress from \relationship{person}{next to}{sofa} to \relationship{person}{in front of}{sofa} to finally \relationship{person}{sitting on}{sofa}. Built upon Charades~\cite{charades}, Action Genome provides $476$K object bounding boxes with $1.72$M relationships across $234$K video frames with $157$ action categories.

Most perspectives on action decomposition converge on the prototypical unit of action-object couplets~\cite{zacks2001perceiving,reynolds2007computational,kurby2008segmentation,lillo2014discriminative}. Action-object couplets refer to transitive actions performed on objects (e.g.~``moving a chair'' or ``throwing a ball'') and intransitive self-actions (e.g.~``moving towards the sofa''). Action Genome's dynamic scene graph representations capture both such types of events and as such, represent the prototypical unit. With this representation, we enable the study for tasks such as spatio-temporal scene graph prediction --- a task where we estimate the decomposition of action dynamics given a video. We can even improve existing tasks like action recognition and few-shot action detection by jointly studying how those actions change visual relationships between objects in scene graphs.

To demonstrate the utility of Action Genome's event decomposition, we introduce a method that extends to a state-of-the-art action recognition model~\cite{lfb} by incorporating spatio-temporal scene graphs as feature banks that can be used to both predict the action as well as the objects and relationships involved.  First, we demonstrate that predicting scene graphs can benefit the popular task of action recognition by improving the state-of-the-art on the Charades dataset~\cite{charades} from $42.5\%$ to $44.3\%$ and to $60.3\%$ when using oracle scene graphs. Second, we show that the compositional understanding of actions induces better generalization by showcasing few-shot action recognition experiments, achieving $42.7\%$ mAP using as few as $10$ training examples. Third, we introduce the task of spatio-temporal scene graph prediction and benchmark existing scene graph models with new evaluation metrics designed specifically for videos.  With a better understanding of the dynamics of human-object interactions via spatio-temporal scene graphs, we aim to inspire a new line of research in more decomposable and generalizable action understanding.

\section{Related work}

We derive inspiration from Cognitive Science, compare our representation with static scene graphs, and survey methods in action recognition and few-shot prediction.

\noindent\textbf{Cognitive science.} Early work in Cognitive Science provides evidence for the regularities with which people identify event boundaries~\cite{michotte2017perception,barker1951one,barker1955midwest}. Remarkably, people consistently, both within and between subjects, carve out video streams into events, actions, and activities~\cite{zacks2001human,casati1996events,hard2006making}. Such findings hint that it is possible to predict when actions begin and end, and have inspired hundreds of Computer Vision datasets, models, and algorithms to study tasks like action recognition~\cite{varol2017long,yue2015beyond,escorcia2016daps,yeung2016end,yeung2018every,karpathy2014large}. Subsequent Cognitive and Neuroscience research, using the same paradigm, has also shown that event categories form partonomies~\cite{newtson1973attribution,zacks2001human,hard2006making}. However, Computer Vision has done little work in explicitly representing the hierarchical structures of actions~\cite{lillo2014discriminative}, even though understanding event partonomies can improve tasks like action recognition.

\noindent\textbf{Action recognition in videos.}
Many research projects have tackled the task of action recognition. A major line of work has focused on developing powerful backbone models to extract useful representations from videos~\cite{c3d,tsn,i3d,slowfast,timeception}. Pre-trained on large-scale databases for action classification~\cite{kinetics700,activitynet}, these backbone models serve as cornerstones for downstream video tasks and action recognition on other datasets. To assist more complicated action understanding, another growing set of research explores structural information in videos including temporal ordering~\cite{trn,lin2019tsm}, object localization~\cite{strg,videotransformer,lfb,baradel2018object,ma2018attend}, and even implicit interactions between objects~\cite{ma2018attend,baradel2018object}. In our work, we contrast against these methods by explicitly using a structured decomposition of actions into objects and relationships.

Table~\ref{tab:dataset_compare} lists some of the most popular datasets used for action recognition. One major trend of video datasets is providing considerably large amount of video clips with single action labels~\cite{activitynet, hacs, kinetics700}. Although these databases have driven the progress of video feature representation for many downstream tasks, the provided annotations treat actions as monolithic events, and do not study how objects and their relationships change during actions/activities. In the mean time, other databases have provided more varieties of annotations: AVA~\cite{ava} localizes the actors of actions, Charades~\cite{charades} contains multiple actions happening at the same time, EPIC-Kitchen~\cite{epickitchen} localizes the interacted objects in ego-centric kitchen videos, DALY~\cite{daly} provides object bounding boxes and upper body poses for $10$ daily activities. Still, scene graph, as a comprehensive structural abstraction of images, has not yet been studied in any large-scale video database as a potential representation for action recognition. In this work, we present Action Genome, the first large-scale database to jointly boost research in scene graphs and action understanding. Compared to existing datasets, we provide orders of magnitude more object and relationship labels grounded in actions.

\noindent\textbf{Scene graph prediction.}
Scene graphs are a formal representation for image information~\cite{johnson2015image,krishna2017visual} in a form of a graph, which is widely used in knowledge bases~\cite{guodong2005exploring,culotta2004dependency,zhou2007tree}. Each scene graph encodes objects as nodes connected together by pairwise relationships as edges. Scene graphs have led to many state of the art models in image captioning~\cite{anderson2016spice}, image retrieval~\cite{johnson2015image,schuster2015generating}, visual question answering~\cite{johnson2017inferring}, relationship modeling~\cite{krishna2018referring}, and image generation~\cite{johnson2018image}. Given its versatile utility, the task of scene graph prediction has resulted in a series of publications~\cite{krishna2017visual,dai2017detecting,liang2017deep,li2017vip,li2017scene,newell2017pixels,xu2017scene,zellers2017neural,yang2018graph,herzig2018mapping} that have explored reinforcement learning~\cite{liang2017deep}, structured prediction~\cite{krahenbuhl2011efficient,desai2011discriminative,tu2010auto}, utilizing object attributes~\cite{farhadi2009describing,parikh2011relative}, sequential prediction~\cite{newell2017pixels}, few-shot prediction~\cite{chen2019scene,dornadula2019visual}, and graph-based~\cite{xu2017scene,li2018factorizable,yang2018graph} approaches. However, all of these approaches have restricted their application to static images and have not modelled visual concepts spatio-temporally.

\noindent\textbf{Few-shot prediction.}
The few-shot literature is broadly divided into two main frameworks. The first strategy learns a classifier for a set of frequent categories and then uses them to learn the few-shot categories~\cite{fe2003bayesian,fei2006one,mishra2018generative}. For example, ZSL uses attributes of actions to enable few-shot~\cite{mishra2018generative}. The second strategy learns invariances or decompositions that enable few-shot classification~\cite{kliper2011one,zhu2018compound,bishay2019tarn,dwivedi2019protogan}. OSS and TARN propose a measurement of similarity or distance measure between video pairs~\cite{kliper2011one,bishay2019tarn}, CMN encodes uses a multi-saliency algorithm to encode videos~\cite{zhu2018compound}, and ProtoGAN creates a prototype vector for each class~\cite{dwivedi2019protogan}. Our framework resembles the first strategy because we use the object and visual relationship representations learned using the frequent actions to identify few-shot actions.

\section{Action Genome}

\begin{figure}[t]
    \centering
    \includegraphics[width=\linewidth]{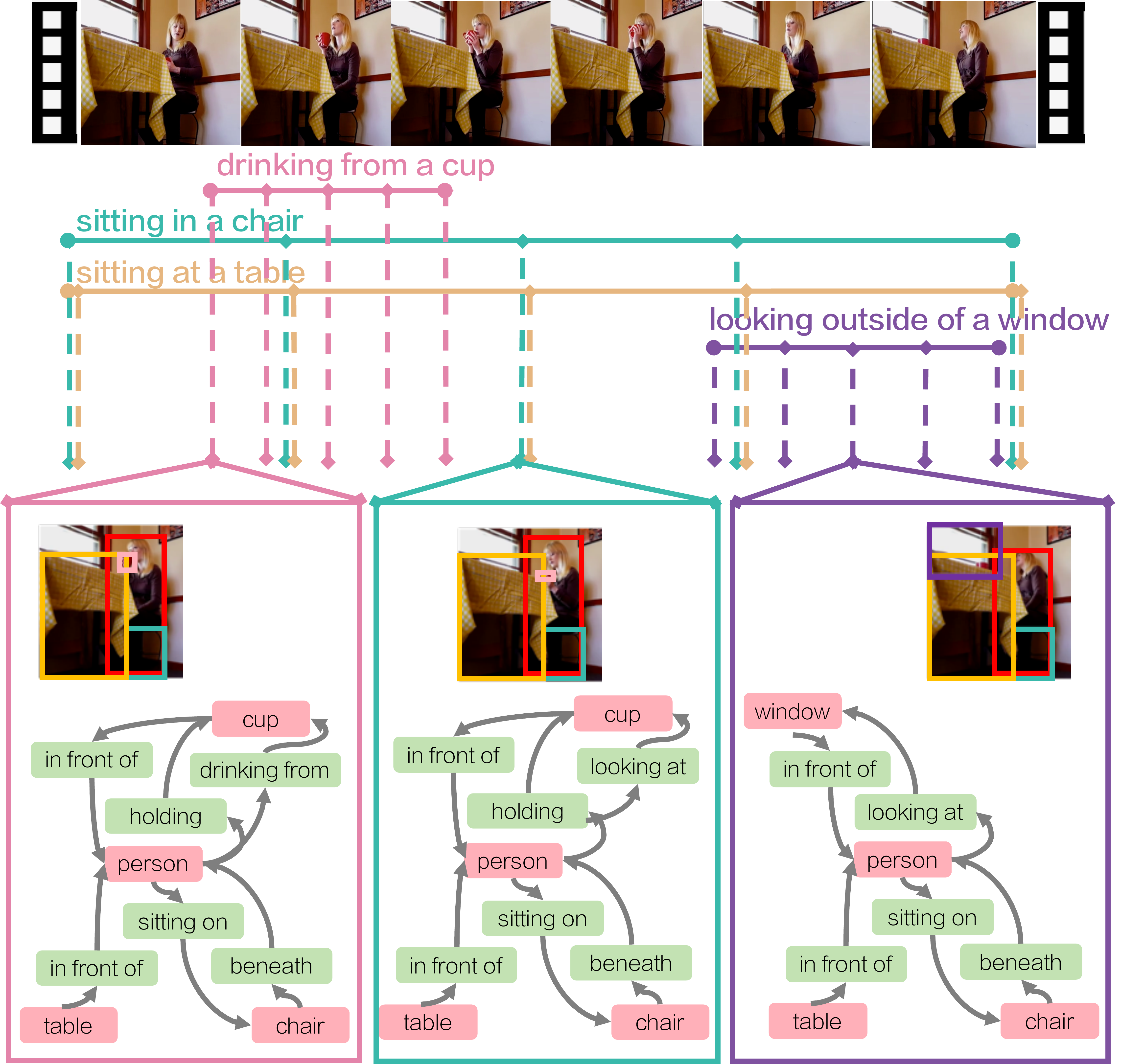}
    \caption{Action Genome's annotation pipeline: For every action, we uniformly sample $5$ frames across the action and annotate the person performing the action along with the objects they interact with. We also annotate the pairwise relationships between the person and those objects. Here, we show a video with $4$ actions labelled, resulting in $20$ ($=4\times5$) frames annotated with scene graphs. The objects are grounded back in the video as bounding boxes.}
    \label{fig:annotation}
\end{figure}

\begin{figure*}[t]
    \centering
    \begin{subfigure}{.4\linewidth}
        \centering
        \includegraphics[width=\linewidth]{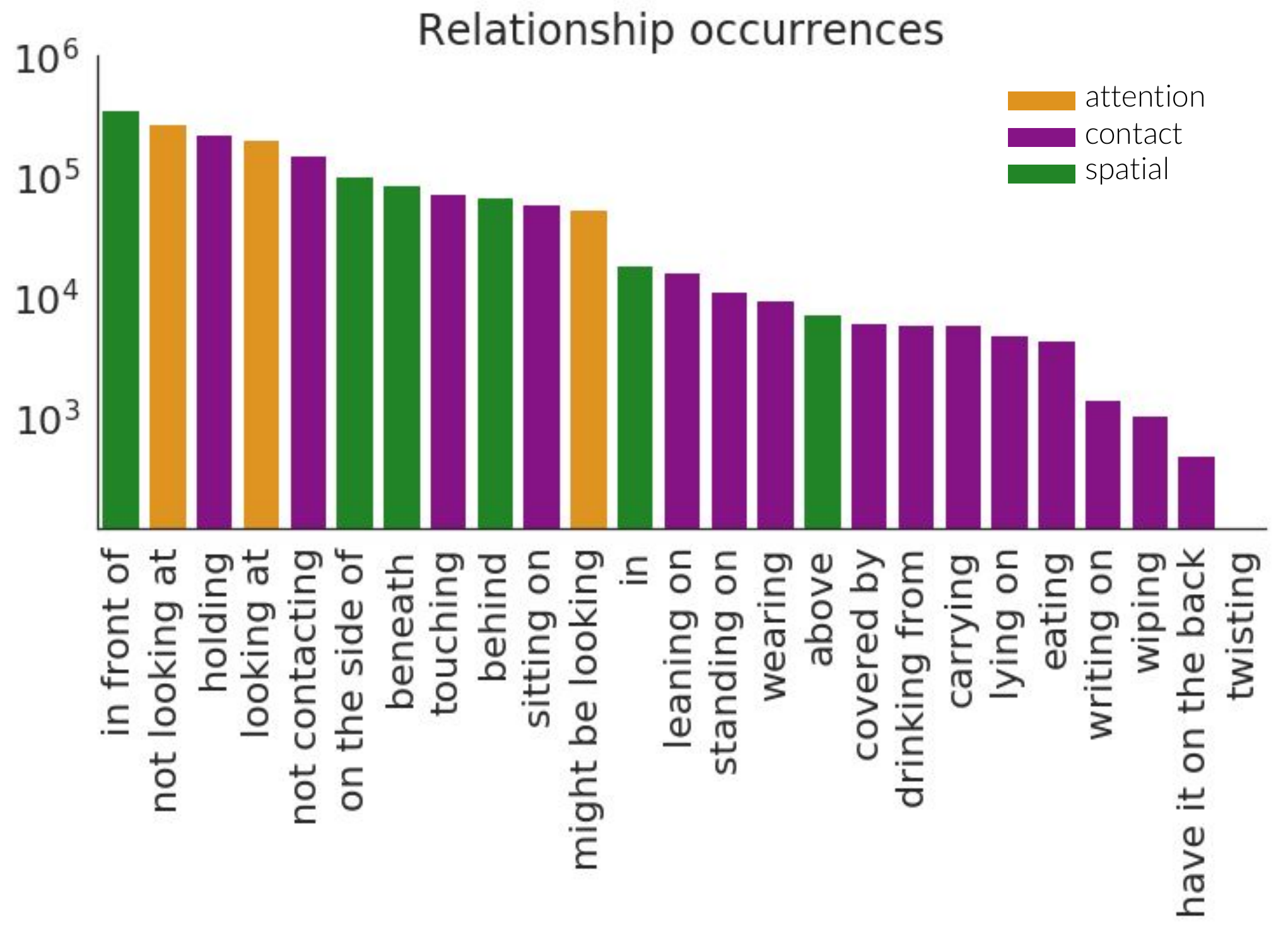}
    \end{subfigure}
    \begin{subfigure}{.5\linewidth}
        \centering
        \includegraphics[width=\linewidth]{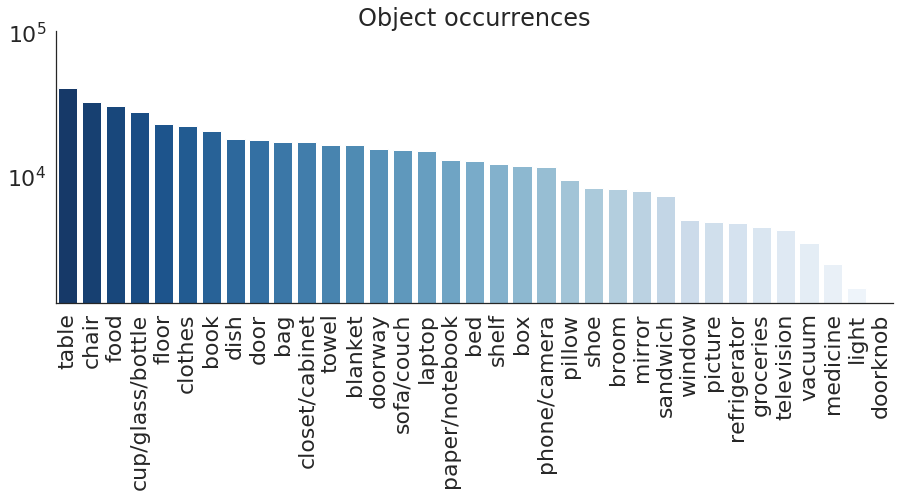}
    \end{subfigure}
    \caption{Distribution of (a) relationship and (b) object occurrences. The relationships are color coded to represent attention, spatial , and contact relationships. Most relationships have at least $1$k instances and objects have at least $10$k instances.}
    \label{fig:occurrences}
\end{figure*}

\begin{table}[t]
\small
\centering
\caption{There are three types of relationships in Action Genome: attention relationships report which objects people are looking at, spatial relationships indicate how objects are laid out spatially, and contact relationships are semantic relationships involving people manipulating objects.}
\resizebox{\linewidth}{!}{%
\begin{tabular}{@{\extracolsep{4pt}}llll@{}}
attention      & spatial          & \multicolumn{2}{c}{contact}   \\
\cline{1-1}\cline{2-2}\cline{3-4}
looking at     & in front of      & carrying            & covered by \\
not looking at & behind           & drinking from       & eating     \\
unsure         & on the side of   & have it on the back & holding    \\
               & above            & leaning on          & lying on   \\
               & beneath          & not contacting      & sitting on \\
               & in               & standing on         & touching   \\
               &                  & twisting            & wearing    \\
               &                  & wiping              & writing on
\end{tabular}}
\label{tab:categories}
\end{table}

\begin{figure}
    \centering
    \includegraphics[width=\columnwidth]{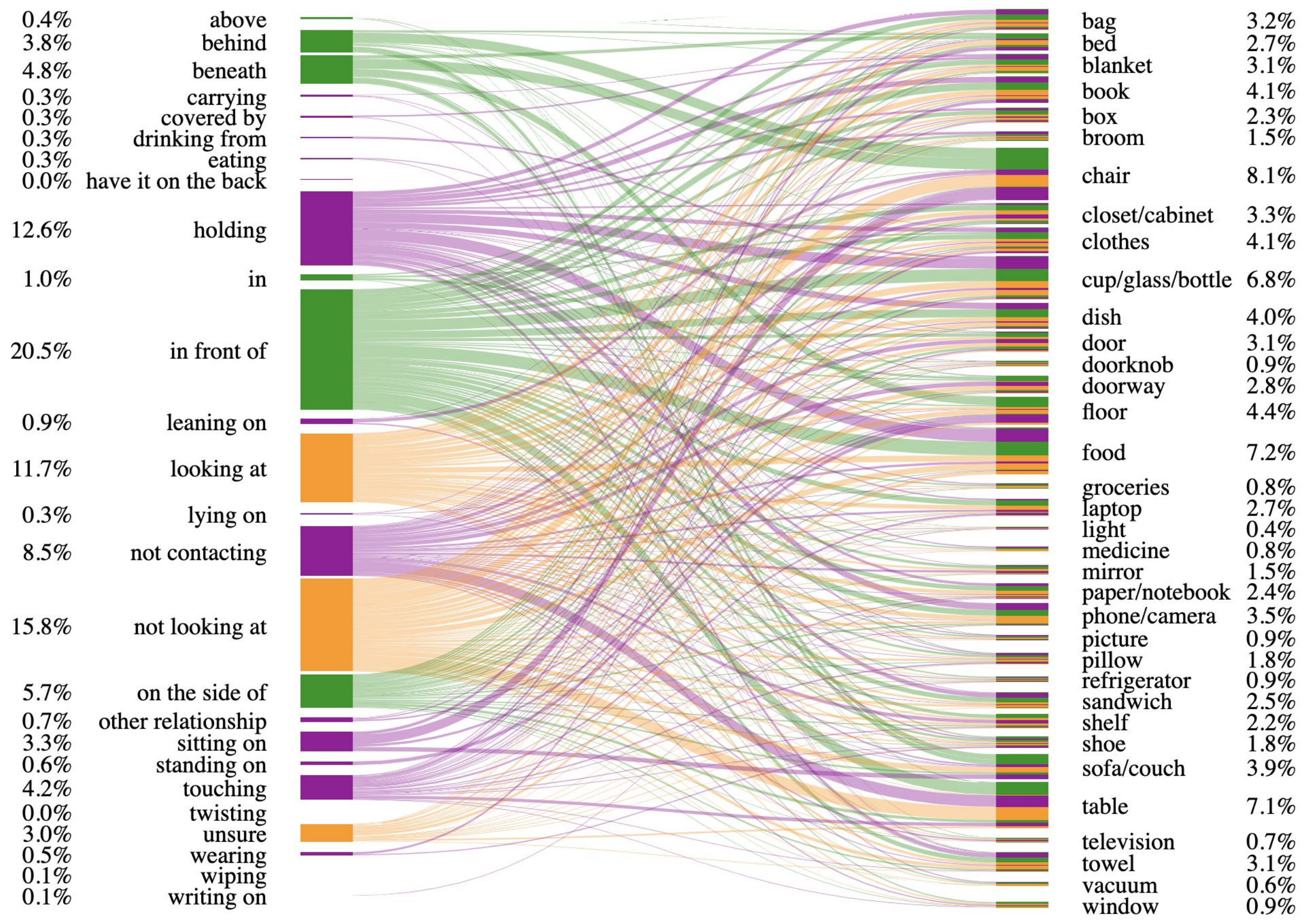}
    \caption{A weighted bipartite mapping between objects and relationships shows that they are densely interconnected in Action Genome. The weights represent percentage of occurrences in which a specific object occurs in a relationship. There are three colors in the graph and they represent the three kinds of relationships: attention (in orange), spatial (in green) and contact (in purple).}
    \label{fig:bipartite_small}
\end{figure}

Inspired from Cognitive Science, we decompose events into prototypical action-object units~\cite{zacks2001perceiving,reynolds2007computational,kurby2008segmentation}. Each action in Action Genome is representated as changes to objects and their pairwise interactions with the actor/person performing the action. We derive our representation as a temporally changing version of Visual Genome's scene graphs~\cite{krishna2017visual}. However, unlike Visual Genome, who's goal was to densely represent a scene with objects and visual relationships, Action Genome's goals is to decompose actions and as such, focuses on annotating only those segments of the video where the action occurs and only those objects that are involved in the action.

\noindent\textbf{Annotation framework.} Action Genome is built upon the Charades dataset~\cite{charades}, which contains $157$ action classes, $144$ of which are human-object activities. In Charades, there are multiple actions that might be occurring at the same time. We do not annotate every single frame in a video; it would be redundant as the changes between objects and relationships occur at longer time scales. 

Figure \ref{fig:annotation} visualizes the pipeline of our annotation. We uniformly sample $5$ frames to annotate across the range of each action interval. With this action-oriented sampling strategy, we provide more labels where more actions occur. For instance, in the example, actions ``sitting on a chair'' and ``drinking from a cup'' occur together and therefore, result in more annotated frames, $5$ from each action. When annotating each sampled frame, the annotators hired were prompted with action labels and clips of the neighboring video frames for context. The annotators first draw bounding boxes around the objects involved in these actions, then choose the relationship labels from the label set. The clips are used to disambiguate between the objects that are actually involved in an action when multiple instances of a given category is present. For example, if multiple ``cups'' are present, the context disambiguates which ``cup'' to annotate for the action ``drinking from a cup''.

Action Genome contains three different kinds of human-object relationships: \textit{attention}, \textit{spatial} and \textit{contact} relationships (see Table \ref{tab:categories}). Attention relationships indicate if a person is looking at an object or not, and serve as indicators for which object the person is or will interacting with. Spatial relationships describe where objects are relative to one another. Contact relationships describe the different ways the person is contacting an object. A change in contact often indicates the occurrence of an actions: for example, changing from \relationship{person}{not contacting}{book} to \relationship{person}{holding}{book} may show an action of ``picking up a book''.

It is worth noting that while Charades provides an injective mapping from each action to a verb, it is different from the relationship labels we provide. Charades' verbs are clip-level labels, such as ``awaken'', while we decompose them into frame-level human-object relationships, such as a sequence of \relationship{person}{lying on}{bed}, \relationship{person}{sitting on}{bed} and \relationship{person}{not contacting}{bed}.

\noindent\textbf{Database statistics.} Action Genome provides frame-level scene graph labels for the components of each action. Overall, we provide annotations for \textbf{$234,253$} frames with a total of \textbf{$476,229$} bounding boxes of $35$ object classes (excluding ``person''), and \textbf{$1,715,568$} instances of $25$ relationship classes. Figure~\ref{fig:occurrences} visualizes the log-distribution of object and relationship categories in the dataset. Like most concepts in vision, some objects (e.g.~\object{table} and \object{chair}) and relationships (e.g.~\predicate{in front of} and \predicate{not looking at}) occur frequently while others (e.g.~\predicate{twisting} and \object{doorknob}) only occur a handful of times. However, even with such a distribution, almost all objects have at least $10$k instances and every relationship as at least $1$K instances.

Additionally, Figure~\ref{fig:bipartite_small} visualizes how frequently objects occur in which relationships. We see that most objects are pretty evenly involved in all three types of relationships. Unlike Visual Genome, where dataset bias provides a strong baseline for predicting relationships given the object categories, Action Genome does not suffer the same bias. 

\section{Method}



\begin{figure}
    \centering
    \includegraphics[width=\linewidth]{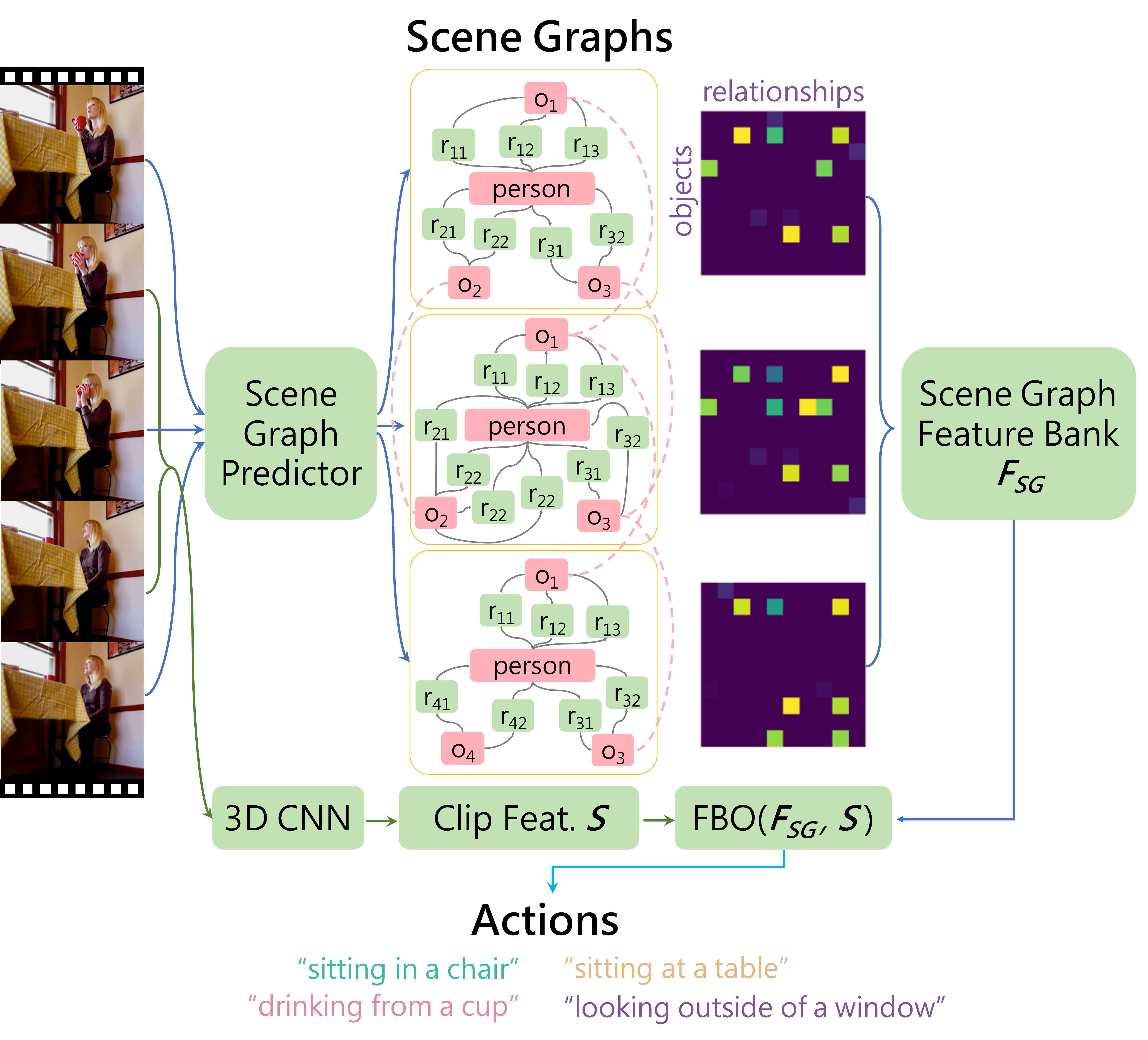}
    \caption{Overview of our proposed model, SGFB, for action recognition using spatio-temporal scene graphs. SGFB predicts scene graphs for every frame in a video. These scene graphs are converted into features representations that are then combined using methods similar to long-term feature banks~\cite{lfb}. The final representation is merged with 3D CNN features and used to predict action labels.}
    \label{fig:arch}
\end{figure}

We validate the utility of Action Genome's action decomposition by studying the effect of simultaneously learning spatio-temporal scene graphs while learning to recognize actions. We propose a method, named Scene Graph Feature Banks (SGFB), to incorporate spatio-temporal scene graphs into action recognition. Our method is inspired by recent work in computer vision that uses the information ``banks''~\cite{li2010object,althoff2012detection,lfb}. Information banks are feature representations that have been used to represent, for example, object categories that occur in the video~\cite{li2010object}, or even include where the objects are~\cite{althoff2012detection}. Our model is most directly related to the recent long-term feature banks~\cite{lfb}, which accumulates features of a long video as a fixed size representation for action recognition. 

Overall, our SGFB model contains two components: the first component generates spatio-temporal scene graphs while the second component encodes the graphs to predict action labels. Given a video sequence $v = \{i_1, i_2, \ldots, i_N\}$, the aim of traditional multi-class action recognition is to assign multiple action labels to this video. Here, $v$ represents the video sequence made up of image frames $i_j \forall j \in [1,N]$. SGFB generates a spatio-temporal scene graph for every frame in the given video sequence.
The scene graphs are encoded to formulate a spatio-temporal scene graph feature bank for the final task of action recognition. We describe the scene graph prediction and the scene graph feature bank components in more detail below. See Figure~\ref{fig:arch} for a high-level visualization of the model's forward pass.

\subsection{Scene graph prediction}
Previous research has proposed plenty of methods for predicting scene graphs on static images~\cite{zellers2017neural,lu2016visual,yang2018graph,li2017scene,xu2017scene,zhang2019graphical}. We employ a state-of-the-art scene graph prediction method as the first step of our method. Given a video sequence $v$, the scene graph predictor $SG$ generates all the objects and connects each object with their relationships with the actor in each frame, i.e. $SG: I \xrightarrow{} G$. On each frame, the scene graph $G = (O, R)$ consists of a set of objects $O = \{o_1, o_2, \dots\}$ that a person is interacting with and a set of relationships $R = \{\{r_{11}, r_{12}, \dots\}, \{r_{21}, r_{22}, \dots\}, \dots\}$. Here $r_{pq}$ denotes the $q-$th relationship between the person with the object $o_p$. Note that there can be multiple relationships between the person and each object, including attention, spatial, and contact relationships. Besides the graph labels, the scene graph predictor $SG$ also outputs confidence scores for all predicted objects: $\{s_{o_1}, s_{o_2},...\}$ and relationships: $\{\{s_{r_{11}}, s_{r_{12}}, \dots\}, \{s_{r_{21}}, s_{r_{22}}, \dots\}, \dots\}$. We have experimented with various choices of $SG$ and benchmark their performance on Action Genome in Section \ref{sec:benchmark}.

\subsection{Scene graph feature banks}
After obtaining the scene graph $G$ on each frame, we formulate a feature vector $f$ by aggregating the information across all the scene graphs into a feature bank. Let's assume there are $\lvert O \rvert$ classes of objects and $\lvert R \rvert$ classes of relationships. In Action Genome,  $\lvert O \rvert =35$ and $\lvert R \rvert =25$. We first construct a confidence matrix $C$ with dimension $\lvert O \rvert \times \lvert R \rvert$, where each entry corresponds to an object-relationship category pair. We compure every entry of this matrix using the scores output by the scene graph predictor $SG$. $C_{ij} = s_{o_i} \times s_{r_{ij}}$. Intuitively, $C_{ij}$ is a high value when $SG$ is confident that there is an object $o_i$ in the current frame and its relationship with the actor is $r_{ij}$. We flatten the confidence matrix as the feature vector $f$ for each image.

Formally, $F_{SG} = [f_1, f_2, ..., f_T]$ is a sequence of scene graph features extracted from frames $i_1, i_2, ..., i_N$. We aggregate the features across the frames using methods similar to long-term feature banks~\cite{lfb}, i.e.~$F_{SG}$ are combined with 3D CNN features $S$ extracted from a short-term clip using feature bank operators (FBO), which can be instantiated as mean/max pooling or non-local blocks~\cite{nl}. The 3D CNN embeds short-term information into $S$ while $F_{SG}$ provides contextual information, critical in modeling the dynamics of complex actions with long time span. The final aggregated feature is then used to predict action labels for the video. 


\section{Experiments}
Action Genome's representation enables us to study few-shot action recognition by decomposing actions into temporally changing visual relationships between objects. It also allows us to benchmark whether understanding the decomposition helps improve performance in action recognition or scene graph prediction individually. To study these benefits afforded by Action Genome, we design three experiments: action recognition, few-shot action recognition, and finally, spatio-temporal scene graph prediction. 

\subsection{Action recognition on Charades} \label{exp:action_recognition}

\begin{table}[]
    \small
    \centering
    \caption{Action recognition on Charades validation set in mAP (\%). We outperform all existing methods when we simultaneously predict scene graphs while performing action recognition. We also find that utilizing ground truth scene graphs can significantly boost performance.}
    \resizebox{\linewidth}{!}{%
    \begin{tabular}{llll}
        Method & Backbone & Pre-train & mAP \\ \hline
        \texttt{I3D + NL} \cite{i3d,nl} & R101-I3D-NL & Kinetics-400 & 37.5 \\
        \texttt{STRG} \cite{strg} & R101-I3D-NL & Kinetics-400 & 39.7 \\
        \texttt{Timeception} \cite{timeception} & R101 & Kinetics-400 & 41.1 \\
        \texttt{SlowFast} \cite{slowfast} & R101 & Kinetics-400 & 42.1 \\
        \texttt{SlowFast+NL} \cite{slowfast,nl} & R101-NL & Kinetics-400 & 42.5 \\
        \texttt{LFB} \cite{lfb} & R101-I3D-NL & Kinetics-400 & 42.5 \\ \hline
        \texttt{SGFB} (ours) & R101-I3D-NL & Kinetics-400 & \textbf{44.3} \\
        \texttt{SGFB Oracle} (ours) & R101-I3D-NL & Kinetics-400 & \textbf{60.3}
    \end{tabular}}
    \label{tab:charades_map}
\end{table} 

We expect that grounding the components that compose an action --- the objects and their relationships --- will improve our ability to predict which actions are occurring in a video sequence. So, we evaluate the utility of Action Genome's scene graphs on the task of action recognition.

\noindent\textbf{Problem formulation.} We specifically study multi-class action recognition on the Charades dataset~\cite{charades}. The Charades dataset contains $9,848$ crowdsourced videos with a length of $30$ seconds on average. At any frame, a person can perform multiple actions out of a nomenclature of $157$ classes. The multi-classification task provides a video sequence as input and expects multiple action labels as output. We train our SGFB model to predict Charades action labels during test time and during training, provide SGFB with spatio-temporal scene graphs as additional supervision.

\noindent\textbf{Baselines.} Previous work has proposed methods for multi-class action recognition and benchmarked on Charades. Recent state-of-the-art methods include applying I3D~\cite{i3d} and non-local blocks~\cite{nl} as video feature extractors (\texttt{I3D+NL}), spatio-temporal region graphs (\texttt{STRG})~\cite{strg}, Timeception convolutional layers (\texttt{Timeception})~\cite{timeception}, SlowFast networks (\texttt{SlowFast})~\cite{slowfast}, and long-term feature banks (\texttt{LFB})~\cite{lfb}. All the baseline methods are pre-trained on Kinetics-400~\cite{kinetics400} and the input modality is RGB.

\noindent\textbf{Implementation details.} \texttt{SGFB} first predicts a scene graph on each frame, then constructs a spatio-temporal scene graph feature bank for action recognition. We use Faster R-CNN~\cite{ren2015faster} with ResNet-101~\cite{resnet} as the backbone for region proposals and object detection. We leverage RelDN~\cite{zhang2019graphical} to predict the visual relationships. Scene graph prediction is trained on Action Genome, where we follow the same train/val splits of videos as the Charades dataset. Action recognition uses the same video feature extractor, hyper-parameters, and solver schedulers as long-term feature banks (\texttt{LFB})~\cite{lfb} for a fair comparison.

\noindent\textbf{Results.}  We report performance of all models using mean average precision (mAP) on Charades validation set in Table~\ref{tab:charades_map}. By replacing the feature banks with spatio-temporal scene graph features, we outperform the state-of-the-art LFB by $1.8\%$ mAP. Our features are smaller in size ($35 \times 25 = 875$ in SGFB versus $2048$ in LFB) but concisely capture the more information for recognizing actions.

We also find that improving object detectors designed for videos can further improve action recognition results. To quantitatively demonstrate the potential of better scene graphs on action recognition, we designed an \texttt{SGFB Oracle} setup. The \texttt{SGFB Oracle} assumes that a perfect scene graph prediction method is available. The spatio-temporal scene graph feature bank therefore, directly encodes a feature vector from ground truth objects and visual relationships for the annotated frames. Feeding such feature banks into the SGFB model, we observe a significant improvement on action recognition: $16\%$ increase on mAP. Such a boost in performance shows the potential of Action Genome and compositional action understanding when video-based scene graph models are utilized to improve scene graph prediction. It is important to note that the performance by \texttt{SGFB Oracle} is not an upper bound on performance since we only utilize ground truth scene graphs for the few frames that have ground truth annotations.

\subsection{Few-shot action recognition}
Intuitively, predicting actions should be easier from a symbolic embedding of scene graphs than from pixels. When trained with very few examples, compositional action understanding with additional knowledge of scene graphs should outperform methods that treat actions as monolithic concept.
We showcase the capability and potential of spatio-temporal scene graphs to generalize to rare actions.

\begin{table}[]
\centering
\caption{Few-shot experiments. With the ability of compositional action understanding, our SGFB demonstrates better generalizability than LFB. The SGFB oracle shows the great potential of how much the scene graph representation could benefit action recognition.}
\begin{tabular}{lccc}
 & 1-shot & 5-shot & 10-shot \\ \hline
\texttt{LFB} \cite{lfb} & 28.3 & 36.3 & 39.6 \\
\texttt{SGFB} (ours) & \textbf{28.8} & \textbf{37.9} & \textbf{42.7} \\
\texttt{SGFB oracle} (ours) & \textbf{30.4} & \textbf{40.2} & \textbf{50.5}
\end{tabular}
\label{tab:few-shot}
\end{table}

\noindent\textbf{Problem formulation.} In our few-shot action recognition experiments on Charades, we split the $157$ action classes into a base set of $137$ classes and a novel set of $20$ classes. We first train a backbone feature extractor (R101-I3D-NL) on all video examples of the base classes, which is shared by the baseline LFB, our SGFB, and SGFB oracle. Next, we train each model with only $k$ examples from each novel class, where $k=1, 5, 10$, for 50 epochs. Finally, we evaluate the trained models on all examples of novel classes in the Charades validation set.



\begin{figure}[t]
    \centering
    \includegraphics[width=\linewidth]{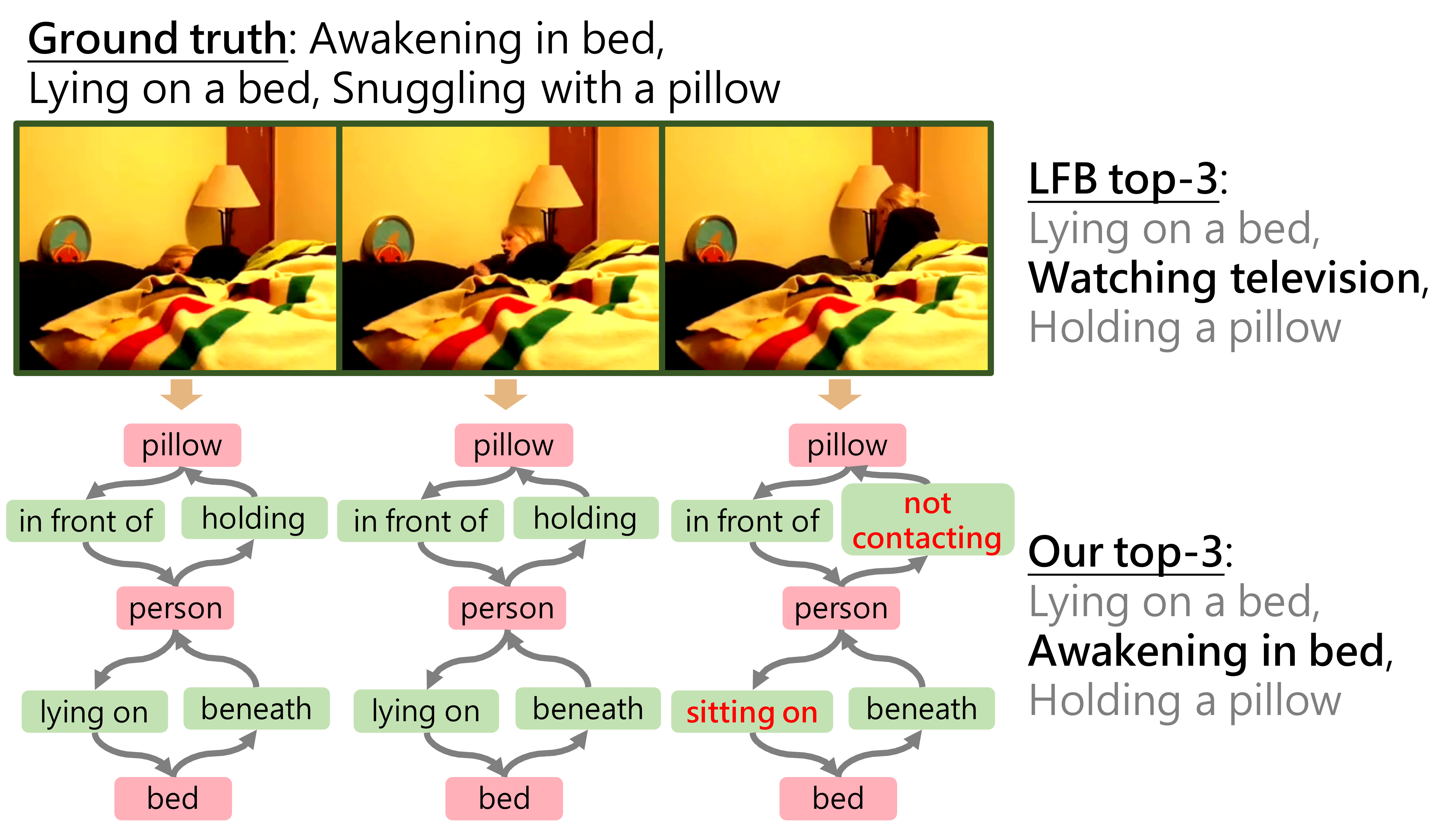}
    \caption{Qualitative results of $10$-shot experiments. We compare the predictions of our \texttt{SGFB} against \texttt{LFB}~\cite{lfb}. Since \texttt{SGFB} uses scene graph knowledge and explicitly captures the dynamics of human-object relationships, it easily learns the concept of ``awakening in bed'' even when only trained with $10$ examples of this label. Also, since \texttt{SGFB} is trained to detect and ground objects, it avoids misclassifying objects, such as \object{television}, which then results in more robust action recognition.}
    \label{fig:qual}
\end{figure}

\noindent\textbf{Results.} We report few-shot experiment performance in Table \ref{tab:few-shot}. \texttt{SGFB} achieves better performance than \texttt{LFB} on all $1,5,10$-shot experiments. Furthermore, if with ground truth scene graphs, \texttt{SGFB Oracle} shows a $10.9\%$ $10$-shot mAP improvement. We visualize the comparison between \texttt{SGFB} and \texttt{LFB} in Figure \ref{fig:qual}. With the knowledge of spatio-temporal scene graphs, \texttt{SGFB} better captures action concepts involving the dynamics of objects and relationships.



\subsection{Spatio-temporal scene graph prediction}
\label{sec:benchmark}
Progress in image-based scene graph prediction has cascaded to improvements across multiple Computer Vision tasks, including image captioning~\cite{anderson2016spice}, image retrieval~\cite{johnson2015image,schuster2015generating}, visual question answering~\cite{johnson2017inferring}, relationship modeling~\cite{krishna2018referring} and image generation~\cite{johnson2018image}. In order to promote similar progress in video-based tasks, we introduce the complementary of spatio-temporal scene graph prediction. Unlike image-based scene graph prediction, which only has a single image as input, this task expects a video as input and therefore, can utilize temporal information from neighboring frames to strength its predictions. In this section, we define the task, its evaluation metrics and report benchmarked results from numerous recently proposed image-based scene graph models applied to this new task.

\begin{table*}[]
\centering
\caption{We evaluate numerous recently proposed image-based scene graph prediction models and provide a benchmark for the new task of spatio-temporal scene graph prediction. We find that there is significant room for improvement, especially since these existing methods were  designed to be conditioned on a single frame and do not consider the entire video sequence as a whole.}
\resizebox{\linewidth}{!}{%
\begin{tabular}{@{\extracolsep{4pt}}lllllllllllll@{}}
\multicolumn{1}{c}{\multirow{3}{*}{Method}} & \multicolumn{4}{c}{\texttt{PredCls}} & \multicolumn{4}{c}{\texttt{SGCls}} & \multicolumn{4}{c}{\texttt{SGGen}} \\ \cline{2-5}\cline{6-9}\cline{10-13} 
\multicolumn{1}{c}{} & \multicolumn{2}{c}{image} & \multicolumn{2}{c}{video} & \multicolumn{2}{c}{image} & \multicolumn{2}{c}{video} & \multicolumn{2}{c}{image} & \multicolumn{2}{c}{video} \\ \cline{2-3} \cline{4-5} \cline{6-7} \cline{8-9} \cline{10-11} \cline{12-13}
\multicolumn{1}{c}{} & \multicolumn{1}{c}{R@20} & \multicolumn{1}{c}{R@50} & \multicolumn{1}{c}{R@20} & \multicolumn{1}{c}{R@50} & \multicolumn{1}{c}{R@20} & \multicolumn{1}{c}{R@50} & \multicolumn{1}{c}{R@20} & \multicolumn{1}{c}{R@50} & \multicolumn{1}{c}{R@20} & \multicolumn{1}{c}{R@50} & \multicolumn{1}{c}{R@20} & \multicolumn{1}{c}{R@50}\\ \hline \hline
\texttt{VRD}~\cite{lu2016visual} & 24.92 & 25.20 & 24.63 & 24.87 & 22.59 & 24.62 & 22.25 & 24.30 & 16.40 & 17.27 & 16.01 & 16.87 \\
\texttt{Freq Prior}~\cite{zellers2017neural} & 45.50 & 45.67 & 44.91 & 45.05 & 43.89 & 45.60 & 43.30 & 44.99 & 33.37 & 34.54 & 32.65 & 33.79 \\
\texttt{Graph R-CNN}~\cite{yang2018graph} & 23.71 & 23.91 & 23.42 & 23.60 & 21.76 & 23.50 & 21.41 & 23.18 & 15.99 & 16.81 & 15.59 & 16.42 \\
\texttt{MSDN}~\cite{li2017scene} & 48.05 & 48.32 & 47.43 & 47.67 & 44.69 & 47.85 & 43.99 & 47.17 & 34.18 & 36.09 & 33.45 & 35.28 \\
\texttt{IMP}~\cite{xu2017scene} & 48.20 & 48.48 & 47.58 & 47.83 & 44.78 & 48.00 & 44.12 & 47.35 & 34.24 & 36.18 & 33.58 & 35.47 \\
\texttt{RelDN}~\cite{zhang2019graphical} & \textbf{49.37} & \textbf{49.58} & \textbf{48.80} & \textbf{48.98} & \textbf{46.74} & \textbf{49.36} & \textbf{46.19} & \textbf{48.76} & \textbf{35.61} & \textbf{37.21} & \textbf{34.92} & \textbf{36.49}
\end{tabular}}
\label{tab:benchmark}
\end{table*}

\noindent\textbf{Problem formulation.} The task expects as input a video sequence $v = \{i_1, i_2, \ldots i_n\}$ where $i_j \forall j \in [1, n]$ represents image frames from the video. The task requires the model to generate a spatio-temporal scene graph $G=(O, R)$ per frame. $o_k \in O$ is represented as objects with category labels and bounding box locations. $r_{j,kl} \in R$ represents the relationships between objects $o_i$ and $o_j$.

\noindent\textbf{Evaluation metrics.} We borrow the three standard evaluation modes for image-based scene graph prediction~\cite{lu2016visual}: (i) scene graph detection (\texttt{SGDET}) which expects input images and predicts bounding box locations, object categories, and predicate labels, (ii) scene graph classification (\texttt{SGCLS}) which expects ground truth boxes and predicts object categories and predicate labels, and (iii) predicate classification (\texttt{PREDCLS}), which expects ground truth bounding boxes and object categories to predict predicate labels. We refer the reader to the paper that introduced these tasks for more details~\cite{lu2016visual}. We adapt these metrics for video, where the per-frame measurements are first averaged in each video as the measurement of the video, then we average video results as the final result for the test set.

\noindent\textbf{Baselines.} We benchmark the following recently proposed image-based scene graph models for the task of spatio-temporal scene graph prediction: VRD's visual module (\texttt{VRD})~\cite{lu2016visual}, iterative message passing (\texttt{IMP})~\cite{xu2017scene}, multi-level scene description network (\texttt{MSDN})~\cite{li2017scene}, graph convolution R-CNN (\texttt{Graph R-CNN})~\cite{yang2018graph}, neural motif's frequency prior (\texttt{Freq-prior})~\cite{zellers2017neural}, and relationship detection network (\texttt{RelDN})~\cite{zhang2019graphical}.

\noindent\textbf{Results.} To our surprise, we find that \texttt{IMP}, which was one of the earliest scene graph prediction models actually outperforms numerous more recently proposed methods. The most recently proposed scene graph model, \texttt{RelDN} marginally outperforms \texttt{IMP}, suggesting that modeling similarlities between object and relationship classes improve performance in our task as well. The small gap in performance between the task of \texttt{PredCls} and \texttt{SGCls} suggests that these models suffer from not being able to accurately detect the objects in the video frames. Improving object detectors designed specifically for videos could improve performance. The models were trained only using Action Genome's data and not finetuned on Visual Genome~\cite{krishna2017visual}, which contains image-based scene graphs, or on ActivityNet Captions~\cite{krishna2017dense}, which contains dense captioning of actions in videos with natural language paragraphs. We expect that finetuning models with such datasets would result in further improvements.

\section{Future work}
With the rich hierarchy of events, Action Genome not only enables research on spatio-temporal scene graph prediction and compositional action recognition, but also promises various research directions. We hope future work will develop methods for the following:

\noindent\textbf{Spatio-temporal action localization.} The majority of spatio-temporal action localization methods~\cite{girdhar2018better,videotransformer,sun2018actor,jiang2018human,sun2018actor} focus on localizing the person performing the action but ignore the objects, which are also involved in the action, that the person interacts with. Action Genome can enable research on localization of both actors and objects, formulating a more comprehensive grounded action localization task. Furthermore, other variants of this task can also be explored; for example, a weakly-supervised localization task where a model is trained with only action labels but tasked with localizing the actors and objects.

\noindent\textbf{Explainable action models.} Explainable visual models is an emerging field of research. Amongst numerous techniques, saliency prediction has emerged as a key mechanism to interpret machine learning models~\cite{simonyan2013deep,mahendran2016salient,selvaraju2017grad}. Action Genome provides frame-level labels of attention in the form of objects that a the person performing the action is either \texttt{looking at} or interacting with. These labels can be used to further train explainable models.

\noindent\textbf{Video generation from spatio-temporal scene graphs.} Recent studies have explored image generation from scene graphs~\cite{johnson2018image,ashual2019specifying}. Similarly, with a structured video representation, Action Genome enables research on video generation from spatio-temporal scene graphs.


\section{Conclusion}
We introduce Action Genome, a representation that decomposes actions into spatio-temporal scene graphs. Scene graphs explain how objects and their relationships change as an action occurs. We demonstrated the utility of Action Genome by collecting a large dataset of spatio-temporal scene graphs and used it to improve state of the art results for action recognition as well as few-shot action recognition. Finally, we  benchmarked results for the new task of spatio-temporal scene graph prediction. We hope that Action Genome will inspire a new line of research in more decomposable and generalizable video understanding.

\noindent\textbf{Acknowledgement.} This work has been supported by
Panasonic. This article solely reflects the opinions and conclusions of its authors and not Panasonic or any entity associated with
Panasonic.

{\small
\bibliographystyle{ieee_fullname}
\bibliography{references}

\begin{thebibliography}{10}\itemsep=-1pt

\bibitem{althoff2012detection}
Tim Althoff, Hyun~Oh Song, and Trevor Darrell.
\newblock Detection bank: an object detection based video representation for
  multimedia event recognition.
\newblock In {\em Proceedings of the 20th ACM international conference on
  Multimedia}, pages 1065--1068. ACM, 2012.

\bibitem{anderson2016spice}
Peter Anderson, Basura Fernando, Mark Johnson, and Stephen Gould.
\newblock Spice: Semantic propositional image caption evaluation.
\newblock In {\em European Conference on Computer Vision}, pages 382--398.
  Springer, 2016.

\bibitem{ashual2019specifying}
Oron Ashual and Lior Wolf.
\newblock Specifying object attributes and relations in interactive scene
  generation.
\newblock In {\em Proceedings of the IEEE International Conference on Computer
  Vision}, pages 4561--4569, 2019.

\bibitem{baradel2018object}
Fabien Baradel, Natalia Neverova, Christian Wolf, Julien Mille, and Greg Mori.
\newblock Object level visual reasoning in videos.
\newblock In {\em Proceedings of the European Conference on Computer Vision
  (ECCV)}, pages 105--121, 2018.

\bibitem{barker1951one}
Roger~G Barker and Herbert~F Wright.
\newblock One boy's day; a specimen record of behavior.
\newblock 1951.

\bibitem{barker1955midwest}
Roger~G Barker and Herbert~F Wright.
\newblock Midwest and its children: The psychological ecology of an american
  town.
\newblock 1955.

\bibitem{bishay2019tarn}
Mina Bishay, Georgios Zoumpourlis, and Ioannis Patras.
\newblock Tarn: Temporal attentive relation network for few-shot and zero-shot
  action recognition.
\newblock {\em arXiv preprint arXiv:1907.09021}, 2019.

\bibitem{activitynet}
Fabian Caba~Heilbron, Victor Escorcia, Bernard Ghanem, and Juan Carlos~Niebles.
\newblock Activitynet: A large-scale video benchmark for human activity
  understanding.
\newblock In {\em Proceedings of the IEEE Conference on Computer Vision and
  Pattern Recognition}, pages 961--970, 2015.

\bibitem{kinetics700}
Joao Carreira, Eric Noland, Chloe Hillier, and Andrew Zisserman.
\newblock A short note on the kinetics-700 human action dataset.
\newblock {\em arXiv preprint arXiv:1907.06987}, 2019.

\bibitem{i3d}
Joao Carreira and Andrew Zisserman.
\newblock Quo vadis, action recognition? a new model and the kinetics dataset.
\newblock In {\em proceedings of the IEEE Conference on Computer Vision and
  Pattern Recognition}, pages 6299--6308, 2017.

\bibitem{casati1996events}
Roberto Casati and A Varzi.
\newblock Events, volume 15 of the international research library of
  philosophy, 1996.

\bibitem{chen2019scene}
Vincent~S Chen, Paroma Varma, Ranjay Krishna, Michael Bernstein, Christopher
  Re, and Li Fei-Fei.
\newblock Scene graph prediction with limited labels.
\newblock {\em arXiv preprint arXiv:1904.11622}, 2019.

\bibitem{culotta2004dependency}
Aron Culotta and Jeffrey Sorensen.
\newblock Dependency tree kernels for relation extraction.
\newblock In {\em Proceedings of the 42nd annual meeting on association for
  computational linguistics}, page 423. Association for Computational
  Linguistics, 2004.

\bibitem{dai2017detecting}
Bo Dai, Yuqi Zhang, and Dahua Lin.
\newblock Detecting visual relationships with deep relational networks.
\newblock In {\em 2017 IEEE Conference on Computer Vision and Pattern
  Recognition (CVPR)}, pages 3298--3308. IEEE, 2017.

\bibitem{epickitchen}
Dima Damen, Hazel Doughty, Giovanni Maria~Farinella, Sanja Fidler, Antonino
  Furnari, Evangelos Kazakos, Davide Moltisanti, Jonathan Munro, Toby Perrett,
  Will Price, et~al.
\newblock Scaling egocentric vision: The epic-kitchens dataset.
\newblock In {\em Proceedings of the European Conference on Computer Vision
  (ECCV)}, pages 720--736, 2018.

\bibitem{desai2011discriminative}
Chaitanya Desai, Deva Ramanan, and Charless~C Fowlkes.
\newblock Discriminative models for multi-class object layout.
\newblock {\em International journal of computer vision}, 95(1):1--12, 2011.

\bibitem{dornadula2019visual}
Apoorva Dornadula, Austin Narcomey, Ranjay Krishna, Michael Bernstein, and Li
  Fei-Fei.
\newblock Visual relationships as functions: Enabling few-shot scene graph
  prediction.
\newblock {\em arXiv preprint arXiv:1906.04876}, 2019.

\bibitem{dwivedi2019protogan}
Sai~Kumar Dwivedi, Vikram Gupta, Rahul Mitra, Shuaib Ahmed, and Arjun Jain.
\newblock Protogan: Towards few shot learning for action recognition.
\newblock {\em arXiv preprint arXiv:1909.07945}, 2019.

\bibitem{escorcia2016daps}
Victor Escorcia, Fabian~Caba Heilbron, Juan~Carlos Niebles, and Bernard Ghanem.
\newblock Daps: Deep action proposals for action understanding.
\newblock In {\em European Conference on Computer Vision}, pages 768--784.
  Springer, 2016.

\bibitem{farhadi2009describing}
Ali Farhadi, Ian Endres, Derek Hoiem, and David Forsyth.
\newblock Describing objects by their attributes.
\newblock In {\em Computer Vision and Pattern Recognition, 2009. CVPR 2009.
  IEEE Conference on}, pages 1778--1785. IEEE, 2009.

\bibitem{fe2003bayesian}
Li Fe-Fei et~al.
\newblock A bayesian approach to unsupervised one-shot learning of object
  categories.
\newblock In {\em Proceedings Ninth IEEE International Conference on Computer
  Vision}, pages 1134--1141. IEEE, 2003.

\bibitem{fei2006one}
Li Fei-Fei, Rob Fergus, and Pietro Perona.
\newblock One-shot learning of object categories.
\newblock {\em IEEE transactions on pattern analysis and machine intelligence},
  28(4):594--611, 2006.

\bibitem{slowfast}
Christoph Feichtenhofer, Haoqi Fan, Jitendra Malik, and Kaiming He.
\newblock Slowfast networks for video recognition.
\newblock In {\em Proceedings of the IEEE International Conference on Computer
  Vision}, pages 6202--6211, 2019.

\bibitem{girdhar2018better}
Rohit Girdhar, Jo{\~a}o Carreira, Carl Doersch, and Andrew Zisserman.
\newblock A better baseline for ava.
\newblock {\em arXiv preprint arXiv:1807.10066}, 2018.

\bibitem{videotransformer}
Rohit Girdhar, Joao Carreira, Carl Doersch, and Andrew Zisserman.
\newblock Video action transformer network.
\newblock In {\em Proceedings of the IEEE Conference on Computer Vision and
  Pattern Recognition}, pages 244--253, 2019.

\bibitem{ava}
Chunhui Gu, Chen Sun, David~A Ross, Carl Vondrick, Caroline Pantofaru, Yeqing
  Li, Sudheendra Vijayanarasimhan, George Toderici, Susanna Ricco, Rahul
  Sukthankar, et~al.
\newblock Ava: A video dataset of spatio-temporally localized atomic visual
  actions.
\newblock In {\em Proceedings of the IEEE Conference on Computer Vision and
  Pattern Recognition}, pages 6047--6056, 2018.

\bibitem{guodong2005exploring}
Zhou GuoDong, Su Jian, Zhang Jie, and Zhang Min.
\newblock Exploring various knowledge in relation extraction.
\newblock In {\em Proceedings of the 43rd annual meeting on association for
  computational linguistics}, pages 427--434. Association for Computational
  Linguistics, 2005.

\bibitem{hard2006making}
Bridgette~M Hard, Barbara Tversky, and David~S Lang.
\newblock Making sense of abstract events: Building event schemas.
\newblock {\em Memory \& cognition}, 34(6):1221--1235, 2006.

\bibitem{resnet}
Kaiming He, Xiangyu Zhang, Shaoqing Ren, and Jian Sun.
\newblock Deep residual learning for image recognition.
\newblock In {\em Proceedings of the IEEE conference on computer vision and
  pattern recognition}, pages 770--778, 2016.

\bibitem{herzig2018mapping}
Roei Herzig, Moshiko Raboh, Gal Chechik, Jonathan Berant, and Amir Globerson.
\newblock Mapping images to scene graphs with permutation-invariant structured
  prediction.
\newblock In {\em Advances in Neural Information Processing Systems}, pages
  7211--7221, 2018.

\bibitem{timeception}
Noureldien Hussein, Efstratios Gavves, and Arnold~WM Smeulders.
\newblock Timeception for complex action recognition.
\newblock In {\em Proceedings of the IEEE Conference on Computer Vision and
  Pattern Recognition}, pages 254--263, 2019.

\bibitem{jiang2018human}
Jianwen Jiang, Yu Cao, Lin Song, Shiwei Zhang4~Yunkai Li, Ziyao Xu, Qian Wu,
  Chuang Gan, Chi Zhang, and Gang Yu.
\newblock Human centric spatio-temporal action localization.
\newblock In {\em ActivityNet Workshop on CVPR}, 2018.

\bibitem{johnson2018image}
Justin Johnson, Agrim Gupta, and Li Fei-Fei.
\newblock Image generation from scene graphs.
\newblock {\em arXiv preprint arXiv:1804.01622}, 2018.

\bibitem{johnson2017inferring}
Justin Johnson, Bharath Hariharan, Laurens van~der Maaten, Judy Hoffman, Li
  Fei-Fei, C~Lawrence Zitnick, and Ross Girshick.
\newblock Inferring and executing programs for visual reasoning.
\newblock {\em arXiv preprint arXiv:1705.03633}, 2017.

\bibitem{johnson2015image}
Justin Johnson, Ranjay Krishna, Michael Stark, Li-Jia Li, David Shamma, Michael
  Bernstein, and Li Fei-Fei.
\newblock Image retrieval using scene graphs.
\newblock In {\em Proceedings of the IEEE conference on computer vision and
  pattern recognition}, pages 3668--3678, 2015.

\bibitem{karpathy2014large}
Andrej Karpathy, George Toderici, Sanketh Shetty, Thomas Leung, Rahul
  Sukthankar, and Li Fei-Fei.
\newblock Large-scale video classification with convolutional neural networks.
\newblock In {\em Proceedings of the IEEE conference on Computer Vision and
  Pattern Recognition}, pages 1725--1732, 2014.

\bibitem{kinetics400}
Will Kay, Joao Carreira, Karen Simonyan, Brian Zhang, Chloe Hillier, Sudheendra
  Vijayanarasimhan, Fabio Viola, Tim Green, Trevor Back, Paul Natsev, et~al.
\newblock The kinetics human action video dataset.
\newblock {\em arXiv preprint arXiv:1705.06950}, 2017.

\bibitem{kliper2011one}
Orit Kliper-Gross, Tal Hassner, and Lior Wolf.
\newblock One shot similarity metric learning for action recognition.
\newblock In {\em International Workshop on Similarity-Based Pattern
  Recognition}, pages 31--45. Springer, 2011.

\bibitem{krahenbuhl2011efficient}
Philipp Kr{\"a}henb{\"u}hl and Vladlen Koltun.
\newblock Efficient inference in fully connected crfs with gaussian edge
  potentials.
\newblock In {\em Advances in neural information processing systems}, pages
  109--117, 2011.

\bibitem{krishna2018referring}
Ranjay Krishna, Ines Chami, Michael Bernstein, and Li Fei-Fei.
\newblock Referring relationships.
\newblock In {\em Computer Vision and Pattern Recognition}, 2018.

\bibitem{krishna2017dense}
Ranjay Krishna, Kenji Hata, Frederic Ren, Li Fei-Fei, and Juan Carlos~Niebles.
\newblock Dense-captioning events in videos.
\newblock In {\em Proceedings of the IEEE international conference on computer
  vision}, pages 706--715, 2017.

\bibitem{krishna2017visual}
Ranjay Krishna, Yuke Zhu, Oliver Groth, Justin Johnson, Kenji Hata, Joshua
  Kravitz, Stephanie Chen, Yannis Kalantidis, Li-Jia Li, David~A Shamma, et~al.
\newblock Visual genome: Connecting language and vision using crowdsourced
  dense image annotations.
\newblock {\em International Journal of Computer Vision}, 123(1):32--73, 2017.

\bibitem{kurby2008segmentation}
Christopher~A Kurby and Jeffrey~M Zacks.
\newblock Segmentation in the perception and memory of events.
\newblock {\em Trends in cognitive sciences}, 12(2):72--79, 2008.

\bibitem{li2010object}
Li-Jia Li, Hao Su, Li Fei-Fei, and Eric~P Xing.
\newblock Object bank: A high-level image representation for scene
  classification \& semantic feature sparsification.
\newblock In {\em Advances in neural information processing systems}, pages
  1378--1386, 2010.

\bibitem{li2017vip}
Yikang Li, Wanli Ouyang, Xiaogang Wang, and Xiao'Ou Tang.
\newblock Vip-cnn: Visual phrase guided convolutional neural network.
\newblock In {\em Computer Vision and Pattern Recognition (CVPR), 2017 IEEE
  Conference on}, pages 7244--7253. IEEE, 2017.

\bibitem{li2018factorizable}
Yikang Li, Wanli Ouyang, Bolei Zhou, Jianping Shi, Chao Zhang, and Xiaogang
  Wang.
\newblock Factorizable net: an efficient subgraph-based framework for scene
  graph generation.
\newblock In {\em European Conference on Computer Vision}, pages 346--363.
  Springer, 2018.

\bibitem{li2017scene}
Yikang Li, Wanli Ouyang, Bolei Zhou, Kun Wang, and Xiaogang Wang.
\newblock Scene graph generation from objects, phrases and region captions.
\newblock In {\em Proceedings of the IEEE Conference on Computer Vision and
  Pattern Recognition}, pages 1261--1270, 2017.

\bibitem{liang2017deep}
Xiaodan Liang, Lisa Lee, and Eric~P Xing.
\newblock Deep variation-structured reinforcement learning for visual
  relationship and attribute detection.
\newblock In {\em Computer Vision and Pattern Recognition (CVPR), 2017 IEEE
  Conference on}, pages 4408--4417. IEEE, 2017.

\bibitem{lillo2014discriminative}
Ivan Lillo, Alvaro Soto, and Juan Carlos~Niebles.
\newblock Discriminative hierarchical modeling of spatio-temporally composable
  human activities.
\newblock In {\em Proceedings of the IEEE Conference on Computer Vision and
  Pattern Recognition}, pages 812--819, 2014.

\bibitem{lin2019tsm}
Ji Lin, Chuang Gan, and Song Han.
\newblock Tsm: Temporal shift module for efficient video understanding.
\newblock In {\em Proceedings of the IEEE International Conference on Computer
  Vision}, pages 7083--7093, 2019.

\bibitem{lu2016visual}
Cewu Lu, Ranjay Krishna, Michael Bernstein, and Li Fei-Fei.
\newblock Visual relationship detection with language priors.
\newblock In {\em European Conference on Computer Vision}, pages 852--869.
  Springer, 2016.

\bibitem{ma2018attend}
Chih-Yao Ma, Asim Kadav, Iain Melvin, Zsolt Kira, Ghassan AlRegib, and Hans
  Peter~Graf.
\newblock Attend and interact: Higher-order object interactions for video
  understanding.
\newblock In {\em Proceedings of the IEEE Conference on Computer Vision and
  Pattern Recognition}, pages 6790--6800, 2018.

\bibitem{mahendran2016salient}
Aravindh Mahendran and Andrea Vedaldi.
\newblock Salient deconvolutional networks.
\newblock In {\em European Conference on Computer Vision}, pages 120--135.
  Springer, 2016.

\bibitem{michotte2017perception}
Albert Michotte.
\newblock {\em The perception of causality}.
\newblock Routledge, 1963.

\bibitem{miller1995wordnet}
George~A Miller.
\newblock Wordnet: a lexical database for english.
\newblock {\em Communications of the ACM}, 38(11):39--41, 1995.

\bibitem{miller1976language}
George~A Miller and Philip~N Johnson-Laird.
\newblock {\em Language and perception.}
\newblock Belknap Press, 1976.

\bibitem{mishra2018generative}
Ashish Mishra, Vinay~Kumar Verma, M~Shiva~Krishna Reddy, S Arulkumar, Piyush
  Rai, and Anurag Mittal.
\newblock A generative approach to zero-shot and few-shot action recognition.
\newblock In {\em 2018 IEEE Winter Conference on Applications of Computer
  Vision (WACV)}, pages 372--380. IEEE, 2018.

\bibitem{newell2017pixels}
Alejandro Newell and Jia Deng.
\newblock Pixels to graphs by associative embedding.
\newblock In {\em Advances in Neural Information Processing Systems}, pages
  2168--2177, 2017.

\bibitem{newtson1973attribution}
Darren Newtson.
\newblock Attribution and the unit of perception of ongoing behavior.
\newblock {\em Journal of Personality and Social Psychology}, 28(1):28, 1973.

\bibitem{parikh2011relative}
Devi Parikh and Kristen Grauman.
\newblock Relative attributes.
\newblock In {\em Computer Vision (ICCV), 2011 IEEE International Conference
  on}, pages 503--510. IEEE, 2011.

\bibitem{ren2015faster}
Shaoqing Ren, Kaiming He, Ross Girshick, and Jian Sun.
\newblock Faster r-cnn: Towards real-time object detection with region proposal
  networks.
\newblock In {\em Advances in neural information processing systems}, pages
  91--99, 2015.

\bibitem{reynolds2007computational}
Jeremy~R Reynolds, Jeffrey~M Zacks, and Todd~S Braver.
\newblock A computational model of event segmentation from perceptual
  prediction.
\newblock {\em Cognitive science}, 31(4):613--643, 2007.

\bibitem{schuster2015generating}
Sebastian Schuster, Ranjay Krishna, Angel Chang, Li Fei-Fei, and Christopher~D
  Manning.
\newblock Generating semantically precise scene graphs from textual
  descriptions for improved image retrieval.
\newblock In {\em Proceedings of the fourth workshop on vision and language},
  pages 70--80, 2015.

\bibitem{selvaraju2017grad}
Ramprasaath~R Selvaraju, Michael Cogswell, Abhishek Das, Ramakrishna Vedantam,
  Devi Parikh, and Dhruv Batra.
\newblock Grad-cam: Visual explanations from deep networks via gradient-based
  localization.
\newblock In {\em Proceedings of the IEEE International Conference on Computer
  Vision}, pages 618--626, 2017.

\bibitem{charades}
Gunnar~A Sigurdsson, G{\"u}l Varol, Xiaolong Wang, Ali Farhadi, Ivan Laptev,
  and Abhinav Gupta.
\newblock Hollywood in homes: Crowdsourcing data collection for activity
  understanding.
\newblock In {\em European Conference on Computer Vision}, pages 510--526.
  Springer, 2016.

\bibitem{simonyan2013deep}
Karen Simonyan, Andrea Vedaldi, and Andrew Zisserman.
\newblock Deep inside convolutional networks: Visualising image classification
  models and saliency maps.
\newblock {\em arXiv preprint arXiv:1312.6034}, 2013.

\bibitem{sun2018actor}
Chen Sun, Abhinav Shrivastava, Carl Vondrick, Kevin Murphy, Rahul Sukthankar,
  and Cordelia Schmid.
\newblock Actor-centric relation network.
\newblock In {\em Proceedings of the European Conference on Computer Vision
  (ECCV)}, pages 318--334, 2018.

\bibitem{c3d}
Du Tran, Lubomir Bourdev, Rob Fergus, Lorenzo Torresani, and Manohar Paluri.
\newblock Learning spatiotemporal features with 3d convolutional networks.
\newblock In {\em 2015 IEEE International Conference on Computer Vision
  (ICCV)}, pages 4489--4497. IEEE, 2015.

\bibitem{tu2010auto}
Zhuowen Tu and Xiang Bai.
\newblock Auto-context and its application to high-level vision tasks and 3d
  brain image segmentation.
\newblock {\em IEEE Transactions on Pattern Analysis and Machine Intelligence},
  32(10):1744--1757, 2010.

\bibitem{varol2017long}
G{\"u}l Varol, Ivan Laptev, and Cordelia Schmid.
\newblock Long-term temporal convolutions for action recognition.
\newblock {\em IEEE transactions on pattern analysis and machine intelligence},
  40(6):1510--1517, 2017.

\bibitem{tsn}
Limin Wang, Yuanjun Xiong, Zhe Wang, Yu Qiao, Dahua Lin, Xiaoou Tang, and Luc
  Van~Gool.
\newblock Temporal segment networks: Towards good practices for deep action
  recognition.
\newblock In {\em European conference on computer vision}, pages 20--36.
  Springer, 2016.

\bibitem{nl}
Xiaolong Wang, Ross Girshick, Abhinav Gupta, and Kaiming He.
\newblock Non-local neural networks.
\newblock In {\em Proceedings of the IEEE Conference on Computer Vision and
  Pattern Recognition}, pages 7794--7803, 2018.

\bibitem{strg}
Xiaolong Wang and Abhinav Gupta.
\newblock Videos as space-time region graphs.
\newblock In {\em Proceedings of the European Conference on Computer Vision
  (ECCV)}, pages 399--417, 2018.

\bibitem{daly}
Philippe Weinzaepfel, Xavier Martin, and Cordelia Schmid.
\newblock Human action localization with sparse spatial supervision.
\newblock {\em arXiv preprint arXiv:1605.05197}, 2016.

\bibitem{lfb}
Chao-Yuan Wu, Christoph Feichtenhofer, Haoqi Fan, Kaiming He, Philipp
  Krahenbuhl, and Ross Girshick.
\newblock Long-term feature banks for detailed video understanding.
\newblock In {\em Proceedings of the IEEE Conference on Computer Vision and
  Pattern Recognition}, pages 284--293, 2019.

\bibitem{xu2017scene}
Danfei Xu, Yuke Zhu, Christopher~B Choy, and Li Fei-Fei.
\newblock Scene graph generation by iterative message passing.
\newblock In {\em Proceedings of the IEEE Conference on Computer Vision and
  Pattern Recognition}, volume~2, 2017.

\bibitem{yang2018graph}
Jianwei Yang, Jiasen Lu, Stefan Lee, Dhruv Batra, and Devi Parikh.
\newblock Graph r-cnn for scene graph generation.
\newblock {\em arXiv preprint arXiv:1808.00191}, 2018.

\bibitem{yao2007introduction}
Benjamin Yao, Xiong Yang, and Song-Chun Zhu.
\newblock Introduction to a large-scale general purpose ground truth database:
  methodology, annotation tool and benchmarks.
\newblock In {\em International Workshop on Energy Minimization Methods in
  Computer Vision and Pattern Recognition}, pages 169--183. Springer, 2007.

\bibitem{yeung2018every}
Serena Yeung, Olga Russakovsky, Ning Jin, Mykhaylo Andriluka, Greg Mori, and Li
  Fei-Fei.
\newblock Every moment counts: Dense detailed labeling of actions in complex
  videos.
\newblock {\em International Journal of Computer Vision}, 126(2-4):375--389,
  2018.

\bibitem{yeung2016end}
Serena Yeung, Olga Russakovsky, Greg Mori, and Li Fei-Fei.
\newblock End-to-end learning of action detection from frame glimpses in
  videos.
\newblock In {\em Proceedings of the IEEE Conference on Computer Vision and
  Pattern Recognition}, pages 2678--2687, 2016.

\bibitem{yue2015beyond}
Joe Yue-Hei~Ng, Matthew Hausknecht, Sudheendra Vijayanarasimhan, Oriol Vinyals,
  Rajat Monga, and George Toderici.
\newblock Beyond short snippets: Deep networks for video classification.
\newblock In {\em Proceedings of the IEEE conference on computer vision and
  pattern recognition}, pages 4694--4702, 2015.

\bibitem{zacks2001human}
Jeffrey~M Zacks, Todd~S Braver, Margaret~A Sheridan, David~I Donaldson,
  Abraham~Z Snyder, John~M Ollinger, Randy~L Buckner, and Marcus~E Raichle.
\newblock Human brain activity time-locked to perceptual event boundaries.
\newblock {\em Nature neuroscience}, 4(6):651, 2001.

\bibitem{zacks2001perceiving}
Jeffrey~M Zacks, Barbara Tversky, and Gowri Iyer.
\newblock Perceiving, remembering, and communicating structure in events.
\newblock {\em Journal of experimental psychology: General}, 130(1):29, 2001.

\bibitem{zellers2017neural}
Rowan Zellers, Mark Yatskar, Sam Thomson, and Yejin Choi.
\newblock Neural motifs: Scene graph parsing with global context.
\newblock {\em arXiv preprint arXiv:1711.06640}, 2017.

\bibitem{zhang2019graphical}
Ji Zhang, Kevin~J Shih, Ahmed Elgammal, Andrew Tao, and Bryan Catanzaro.
\newblock Graphical contrastive losses for scene graph parsing.
\newblock In {\em Proceedings of the IEEE Conference on Computer Vision and
  Pattern Recognition}, pages 11535--11543, 2019.

\bibitem{hacs}
Hang Zhao, Antonio Torralba, Lorenzo Torresani, and Zhicheng Yan.
\newblock Hacs: Human action clips and segments dataset for recognition and
  temporal localization.
\newblock In {\em Proceedings of the IEEE International Conference on Computer
  Vision}, pages 8668--8678, 2019.

\bibitem{trn}
Bolei Zhou, Alex Andonian, Aude Oliva, and Antonio Torralba.
\newblock Temporal relational reasoning in videos.
\newblock In {\em Proceedings of the European Conference on Computer Vision
  (ECCV)}, pages 803--818, 2018.

\bibitem{zhou2007tree}
Guodong Zhou, Min Zhang, DongHong Ji, and Qiaoming Zhu.
\newblock Tree kernel-based relation extraction with context-sensitive
  structured parse tree information.
\newblock In {\em Proceedings of the 2007 Joint Conference on Empirical Methods
  in Natural Language Processing and Computational Natural Language Learning
  (EMNLP-CoNLL)}, 2007.

\bibitem{zhu2018compound}
Linchao Zhu and Yi Yang.
\newblock Compound memory networks for few-shot video classification.
\newblock In {\em Proceedings of the European Conference on Computer Vision
  (ECCV)}, pages 751--766, 2018.

\bibitem{evar}
Tao Zhuo, Zhiyong Cheng, Peng Zhang, Yongkang Wong, and Mohan Kankanhalli.
\newblock Explainable video action reasoning via prior knowledge and state
  transitions.
\newblock In {\em Proceedings of the 27th ACM International Conference on
  Multimedia}, pages 521--529. ACM, 2019.

\end{thebibliography}
}


\end{document}